# Adaptive Synthetic Characters for Military Training

**Volkan Ustun, Rajay Kumar, and Adam Reilly**
Institute for Creative Technologies, USC
Playa Vista, CA
ustun@ict.usc.edu, kumar@ict.usc.edu, reilly@ict.usc.edu

**Seyed Sajjadi, Andrew Miller**
nFlux, Inc.
Los Angeles, CA
ssajjadi@nflux.ai, amiller@nflux.ai

## ABSTRACT

Behaviors of the synthetic characters in current military simulations are limited since they are generally generated by rule-based and reactive computational models with minimal intelligence. Such computational models cannot adapt to reflect the experience of the characters, resulting in brittle intelligence for even the most effective behavior models devised via costly and labor-intensive processes. Observation-based behavior model adaptation that leverages machine learning and the experience of synthetic entities in combination with appropriate prior knowledge can address the issues in the existing computational behavior models to create a better training experience in military training simulations. In this paper, we introduce a framework that aims to create autonomous synthetic characters that can perform coherent sequences of believable behavior while being aware of human trainees and their needs within a training simulation. This framework brings together three mutually complementary components. The first component is a Unity-based simulation environment - Rapid Integration and Development Environment (RIDE) - supporting One World Terrain (OWT) models and capable of running and supporting machine learning experiments. The second is Shiva, a novel multi-agent reinforcement and imitation learning framework that can interface with a variety of simulation environments, and that can additionally utilize a variety of learning algorithms. The final component is the Sigma Cognitive Architecture that will augment the behavior models with symbolic and probabilistic reasoning capabilities. We have successfully created proof-of-concept behavior models leveraging this framework on realistic terrain as an essential step towards bringing machine learning into military simulations: (1) in order to improve the quality and complexity of non-player characters in training simulations; (2) in order to create more realistic and challenging training experiences while reducing the cost and time to develop them; and (3) in order to make simulations less dependent on the availability of human participants.

## ABOUT THE AUTHORS

**Volkan Ustun** is a senior artificial intelligence researcher at the USC Institute for Creative Technologies. His general research interests are cognitive architectures, computational cognitive models, natural language and simulation. He is a member of the cognitive architecture group and a major contributor to the Sigma cognitive architecture.

**Rajay Kumar** is a research developer at the USC Institute for Creative Technologies. He integrates artificial intelligence models with realistic environments.

**Adam Reilly** is an ICT Computer Scientist developing the Rapid Integration and Development Environment (RIDE) used in the Synthetic Training Environment (STE). This work focuses on the rapid development, integration, distribution, and scalability of research and production level capabilities, used across the DOD and research communities.

**Seyed Sajjadi** is an artificial intelligence researcher at nFlux AI and a member of the cognitive architecture group at USC ICT. His general research interest lies primarily in the theory and application of artificial intelligence, particularly in developing novel approaches for interactive machine learning and structured prediction.

**Andrew Miller** is a research engineer at nFlux AI. His general research interests are artificial intelligence and neuroscience.

# Adaptive Synthetic Characters for Military Training

**Volkan Ustun, Rajay Kumar, and Adam Reilly**
**Institute for Creative Technologies, USC**
**Playa Vista, CA**
ustun@ict.usc.edu, kumar@ict.usc.edu, reilly@ict.usc.edu

**Seyed Sajjadi, Andrew Miller**
**nFlux, Inc.**
**Los Angeles, CA**
ssajjadi@nflux.ai, amiller@nflux.ai

## INTRODUCTION

New-generation simulated training environments, such as Synthetic Training Environment (STE), demand adaptive human-like autonomous synthetic characters to fill human participants' roles in training scenarios. For instance, providing effective training virtually at the point-of-need could only be possible by having realistic synthetic characters that are capable of performing a variety of roles in different situations and environments, which may be complex, continuous, stochastic, partially-observable, and non-stationary with multiple players, either collaborating or competing against one another. Such synthetic characters need to be adaptive since they are almost impossible to generate by strictly controlling their behavior. This approach would require making rules for each task, each environment, and every possible interaction. Instead, observation-based behavior model adaptation is needed to develop them, which will, in turn, make the training simulations less dependent on the availability of human participants that would otherwise be required in addition to the human trainees.

Multi-agent Reinforcement Learning (MARL) presents opportunities in the challenging task of the computational generation of realistic and strategic intelligent behavior for autonomous synthetic characters. MARL models multiple agents that learn by dynamically interacting with an environment and each other, providing a framework for the evaluation of competitive and collaborative dynamics between these agents. Recent advances in MARL research make these models promising candidates to generate adaptive opponent behaviors. State-of-the-art MARL models blend advances in Artificial Neural Networks with algorithms from MARL research (Buşoniu et al. 2010), yielding excellent results for toy problems like predator-prey, cooperative navigation, and physical deception (Lowe et al. 2017). More recently, applications of deep MARL models have shown great promise in more complex environments, such as soccer simulations (Kurach et al. 2019) and first-person shooters (Jaderberg et al., 2019). However, MARL requires simulation environments capable of running quickly (much faster than real-time) for timely and efficient training of such agent models.

In this paper, we discuss our research framework and proof-of-concept models on observation-based behavior model adaptation for synthetic characters in military training simulations. This framework brings together three mutually complementary components. The first component is Rapid Integration and Development Environment (RIDE), which is a Unity-based military simulation environment utilizing One World Terrain (OWT) that can: (1) run and support ML experiments, and (2) provide trainee-controlled interaction experiences to control and change behavior at different stages of the simulation. The second is Shiva: a novel and adaptable multi-agent reinforcement and imitation learning framework that is capable of interfacing with a variety of simulation environments, and utilizing a set of learning algorithms. The final component is the Sigma cognitive architecture (Rosenbloom, Demski, and Ustun 2016), which combines lessons from symbolic cognitive architectures with probabilistic graphical model and neural models, towards a uniform grand unification by leveraging a generalized representation.

The next two sections introduce and discuss the RIDE simulation environment and the Shiva library. Then, we present our proof-of-concept models that leverage both RIDE and Shiva. Finally, we discuss our current efforts in developing realistic scenarios in RIDE and our plans to integrate the Sigma cognitive architecture into the overall framework.

## RAPID INTEGRATION AND DEVELOPMENT ENVIRONMENT (RIDE)

Rapid Integration & Development Environment (RIDE) is a simulation environment uniting many DoD and Army simulation efforts to provide an accelerated development platform and a rapid prototyping sandbox. RIDE is actively being developed at the USC Institute for Creative Technologies (ICT), and it aims to offer a direct benefit to STE as well as the larger DoD and Army simulation communities.

RIDE integrates a range of capabilities, including One World Terrain, training management tools, Non-Player Characters (NPCs) and their respective behavior models, physics models, multiplayer networking, destructibility, and multi-platform support. The One World Terrain data sets are generated from photogrammetric data captures of real-world military bases, allowing agents to train on the same terrain as military personnel (Figure 1). RIDE's goal is to create a simple, drag-and-drop development environment that is usable by people across all technical levels. It leverages robust game engine technology while designed to be agnostic to any specific game or simulation engine. It also provides modelers and researchers with the tools needed to define requirements better and identify potential solutions in much less time and much-reduced costs.

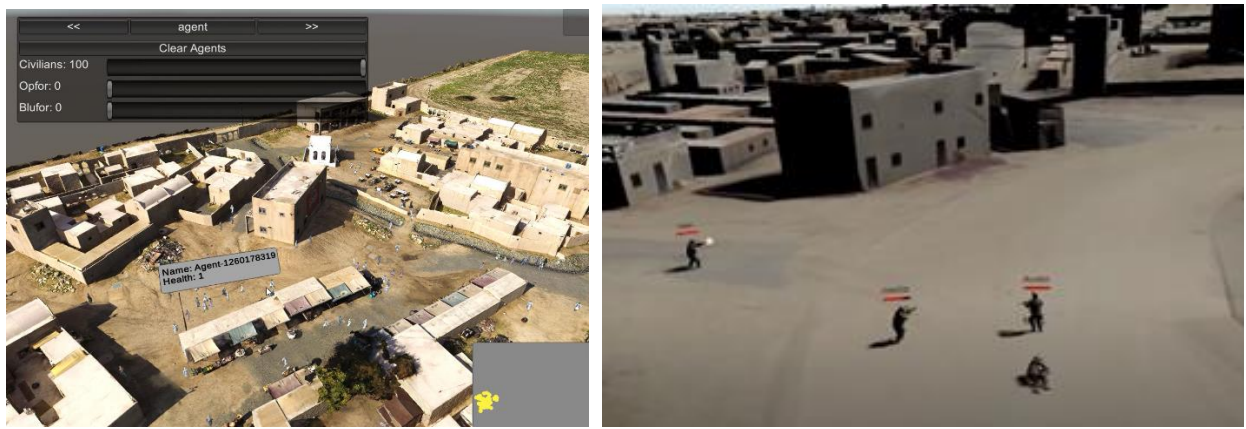

**Figure 1.** RIDE scenes instantiated with OWT data sets.

RIDE currently supports the Unity engine and leverages the ML-Agents Toolkit (Juliani et al. 2018). Developed with continuous feedback from Artificial Intelligence (AI) research groups at ICT, RIDE aspires to adequately support the data, speed, and interfacing needs of state-of-the-art AI models that need interaction with realistic simulation environments for effective training. Timely and efficient training of such AI models requires simulation environments capable of running much faster than real-time while generating and supplying the data that is needed, and RIDE addresses such needs by integrating them into its architectural design. It currently supports creating, training, and testing of AI behaviors utilizing reinforcement learning and imitation learning while using industry-standard open-source Machine Learning (ML) libraries such as PyTorch (Pazske et al., 2019) and TensorFlow (Abadi et al., 2016) as well as the Shiva library, which is introduced in the next section. Since RIDE brings together different capabilities such as One World Terrain data sets and functionalities as well as Army validated Physical Knowledge Acquisition Document (PKAD) models and behaviors, it is possible to provide more accurate representations of the environment and the entities, assisting AI models trained with RIDE to learn more realistic models and representations.

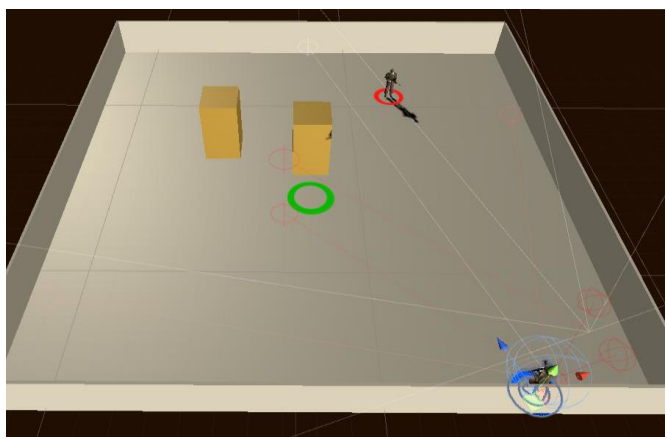

**Figure 2.** Sample take cover setup.

Initial experiences of leveraging RIDE as a fast simulation environment for AI experiments are encouraging, and RIDE is a promising platform for AI research for military training simulations. In an exploratory experiment, it was used to train an agent to "take cover," i.e., find a safe position to stand that obscures enemy fire (Figure 2). Nine agents were simultaneously trained in nine different randomly generated layouts over 15 minutes, which was 500,000 environment steps, to achieve the result of the agent being able to properly take cover in less than 15 seconds the majority of the time. During the training process, the agent used observation vectors (Figure 2) to perceive objects in the environment. All objects were tagged as either "enemy," "cover," or "wall." The agents allowed actions per training step were movement in the cardinal directions and rotation to the left or the right of their current facing orientation. When the agent enters the proper cover area (the green circle), it is rewarded with 5 points. In an effort to force the agent to find the desired position as fast as possible, they receive a small movement penalty (1/5000) every time they execute an action.

After the training was completed, the trained agent was pitted against a scripted agent to gauge the performance of an agent trained with the Proximal Policy Optimization (PPO) algorithm (Schulman et al., 2017) vs. a scripted agent (Figure 3) at a new randomly generated layout. The scripted agent has access to all information about the layout. Hence, it calculates the optimal cover position, which is the closest point that provides full cover from the other agent. On the other hand, the RL agent has access to its own observations and the policy it had learned in training to find a cover location. In this experiment, the RL agent performs competitively and always finds the optimal cover location by utilizing its local observations and the learned policy.

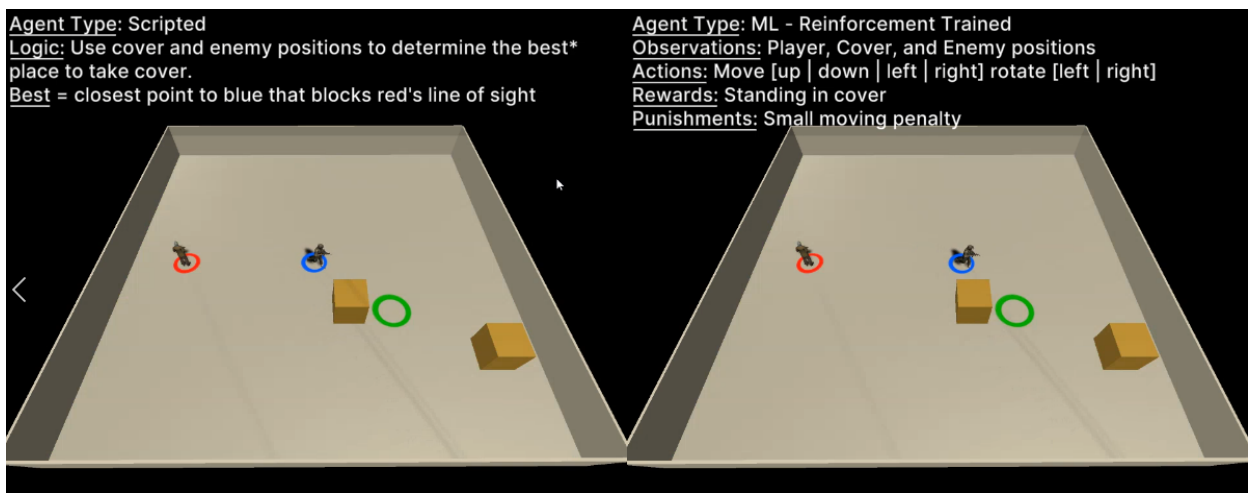

**Figure 3.** Scripted vs. RL trained agents.

**SHIVA**

Multi-agent reinforcement learning (MARL) models multiple agents that learn by dynamically interacting with an environment and each other, providing a framework for the evaluation of competitive and collaborative strategies between these agents. Current state-of-the-art blends advances in deep learning and algorithms from MARL research yielding promising policies that generate adaptive opponent behaviors as well as to effective strategic coordination plans to utilize against potential adversaries. Here, we introduce Shiva: a novel and adaptable multi-agent reinforcement and imitation learning (MARIL) framework that aims to learn proficiency in complex environments efficiently. In the preceding months, Shiva's distributed learning architecture has been actualized and has demonstrated to effectively and efficiently coordinate autonomous agents at scale in cooperative and competitive scenarios under diverse conditions. First and foremost, we have confirmed that Shiva can converge to strategic control decisions in team-based combat simulations with no prior knowledge or guidance about the simulations' rules. These experiments were performed in Unity, MuJoCo, NeuralMMO (Suarez et al., 2019), Open AI Gym, and RoboCup Soccer Simulation 2D engines. Shiva has also proven that it can train agents to imitate human-level strategies in the RoboCup Soccer 2D engine by generating clones of the sophisticated digital soccer teams.

Shiva is built to be simulation-engine agnostic; its framework is abstracted to support various types of observation and action spaces with different environment settings and the number of agents. Additionally, Shiva is designed to support distributed processing across a large number of servers to support learning in complex environments with large observation or action spaces where multiple agents need to converge to a team policy. At the moment, Shiva supports popular reinforcement and imitation learning algorithms such as Deep Q-Network (DQN) (Mnih et al., 2015),

Deep Recurrent Q-Network (DRQN) (Hausknecht and Stone, 2017), Deep Deterministic Policy Gradient (DDPG) (Lillicrap et al, 2019), Proximal Policy Optimizations (PPO) (Schulman et al., 2017), Multi-Agent Deep Deterministic Policy Gradient (MADDPG) (Lowe et al., 2020), Dataset Aggregation (DAGGER) method (Ross, Gordon, and Bagnell 2011) in addition to a few customized and hybrid model-based algorithms that leverage the dynamics of the environment to converge to at a faster rate. The framework is built to enable researchers to design and experiment with new algorithms and be able to test them at scale in different environments and scenarios with minimum setup on the infrastructure.

Shiva leverages imitation learning to bootstrap experts' knowledge to converge to competent strategies in complex environments at a faster rate than vanilla RL algorithms that learn from executing random actions with no prior knowledge until the autonomous agents converge to optimum action policy. Additionally, Shiva optimizes expert's policy through population-based training (Jaderberg et al., 2019) that generates multiple clones of each autonomous agent and matches them against one another to boost the overall intelligence of the agent population which means through self-play, each agent learns the strength and weaknesses of its strategy, resulting in more robust and generalized policies that are learned during each iteration. Shiva's learning paradigm is depicted in Figure 4.

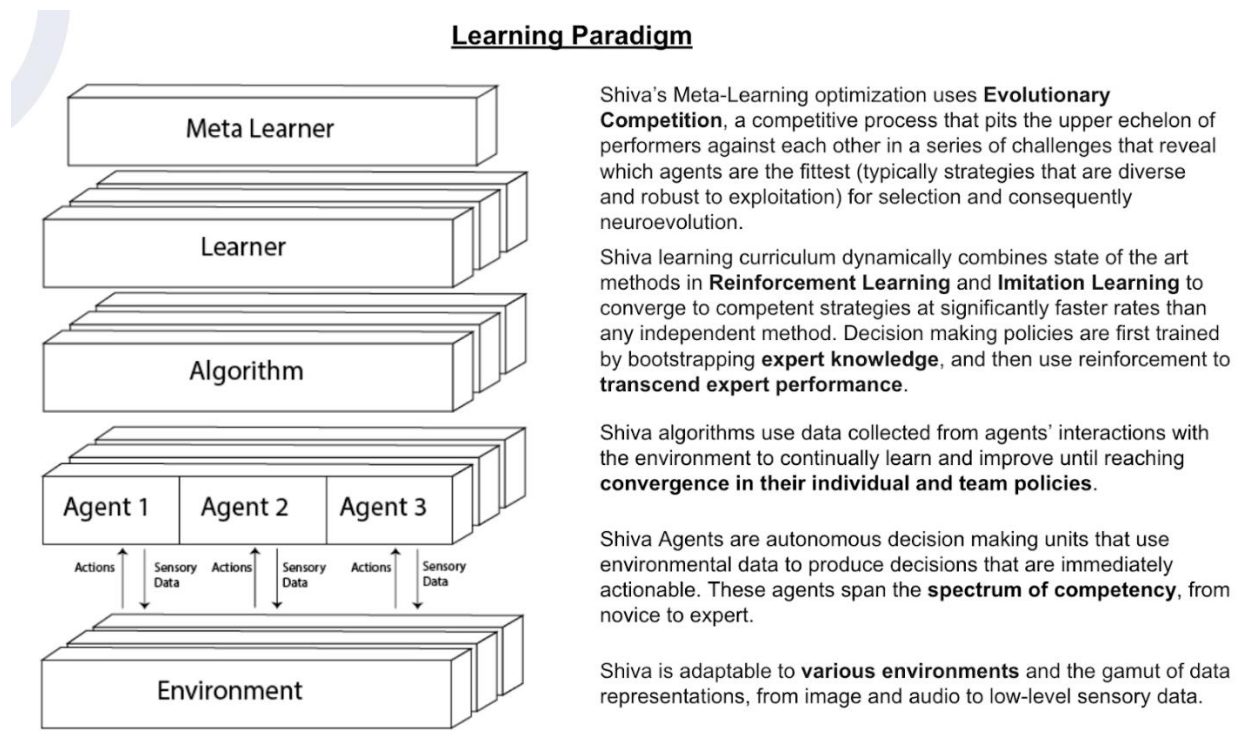

**Figure 4.** Shiva's learning paradigm

Shiva takes advantage of synthetic data generation by modifying the noise in the simulation environment across experiments. Such modification helps with learning more robust strategies to better deal with imperfect information in the task. We leverage these representations in Shiva to create a 'look-ahead' functionality, effectively creating an imagination core that enables each agent to mentally evaluate the outcome of their future actions. Shiva demonstrates resilience in state spaces that contain noisy and imperfect information, by taking advantage of synthetic data generation in a novel approach we refer to as counterfactual policy regularization. Specifically, during training time, we impose noise in the observation space and, in extreme cases, simulate complete sensor failure. In this way, learning agents are forced to avoid being too dependent on particular signals, resulting in agents that are more robust to imperfect information, anomalies, and hardware failures. Figure 5 depicts the architecture of Shiva in detail. Each component is treated as an independent entity, and information gets passed between such components using uniquely identifiable agents and environments. This means as agents learn through the "learner" class or in "meta-learner," a log is kept of the previous policies that can be replicated in test environments at any point in time.

Shiva's architecture is built with modularity in mind; it is agnostic of the environment in which agents learn a task or the framework through which agents perform optimization (Tensforflow or PyTorch). The objective is to provide a unifying framework where researchers and practitioners can test a variety of different algorithms across many different environments without having to construct the underlying scalable architecture needed to learn complex tasks.

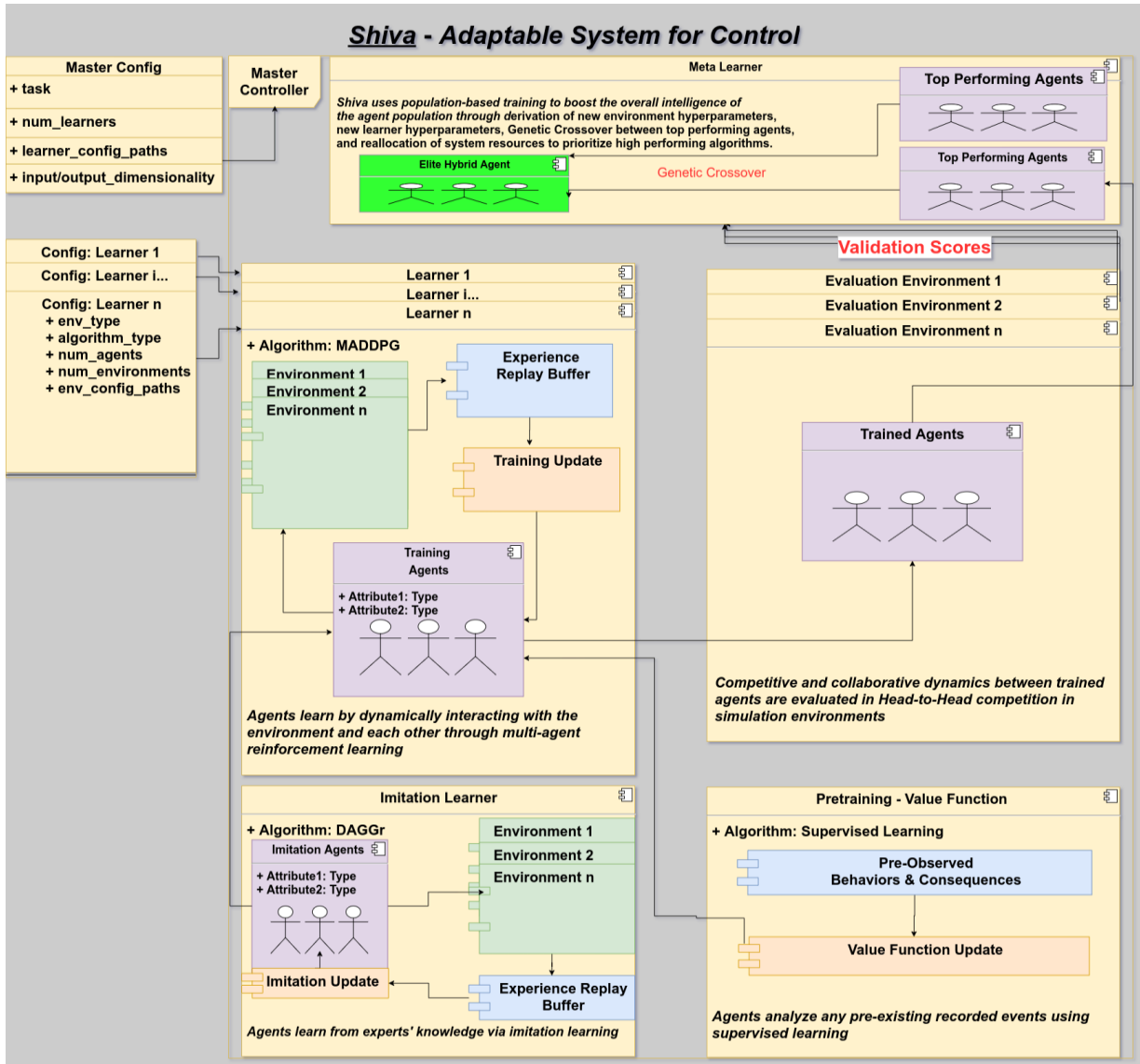

**Figure 5.** Shiva's internal architecture

**PROOF-OF-CONCEPT MODELS**

A set of room clearing scenarios on a grid representation is defined for the proof-of-concept experiments. In these experiments, the agents are split into blue and red teams of varying sizes, and the goal is to eliminate all the members of the opposing team. The behavior of the blue team is determined by a script encoding a heuristic strategy (enter the room one-by-one and then take a position in the center and lower left corners for a three agent team), whereas the red team leverages RL models to learn policies to defeat the blue team. Agents receive rewards for eliminating opposing agents and for the number of teammates that survive a trial. Room clearing scenarios were initially created in NeuralMMO (Suarez et al., 2019), a grid-based environment for training multiple simultaneous agents in a Massively Multiplayer Online Role-Playing Game (MMORPG or MMO). Later, these scenarios are re-created in Unity using character models and environments from RIDE and OWT. The sample grid-based layout used in these scenarios is shown in Figure 6.

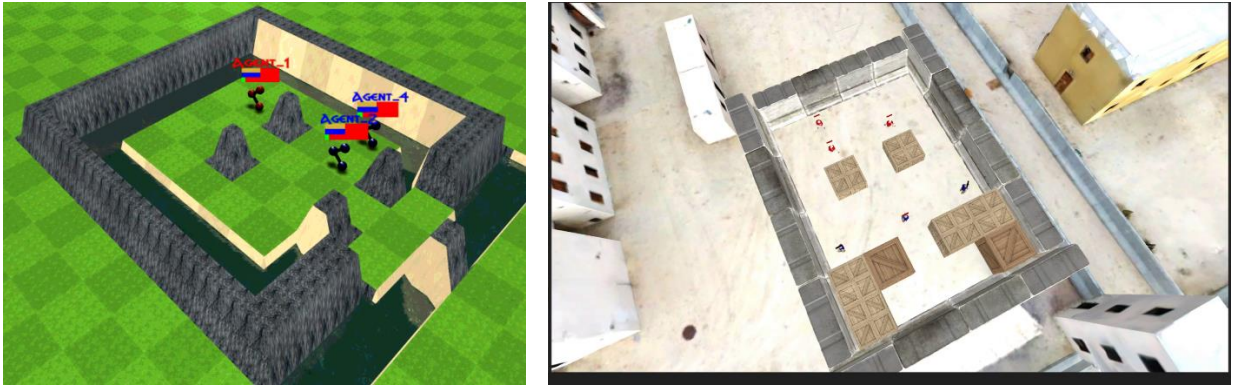

**Figure 6.** Example layout in Neural MMO (on the left) and RIDE (on the right) for the room clearing scenarios

One of the main approaches for multi-agent learning is to leverage the actor-critic concept in multi-agent environments (Konda and Tsitsiklis 2000). For example, DDPG (Lillicrap et al., 2019) is an off-policy algorithm for discrete action spaces., combines the actor-critic concept with Q-learning (Mnih et al., 2015). DDPG makes use of a Q network, referred to as the critic network, a deterministic policy function for the actor, a target Q network, and a target policy network. A replay buffer is also utilized for stability as a source for sampling experiences that are used to update the Q and policy networks. However, multi-agent learning experiments that use independently learning actor-critic pairs are shown to perform poorly in practice. One issue is that each agent's policy changes during training, resulting in a non- stationary environment for each individual actor/agent's perspective. This presents learning stability challenges and prevents the straightforward use of experience replay.

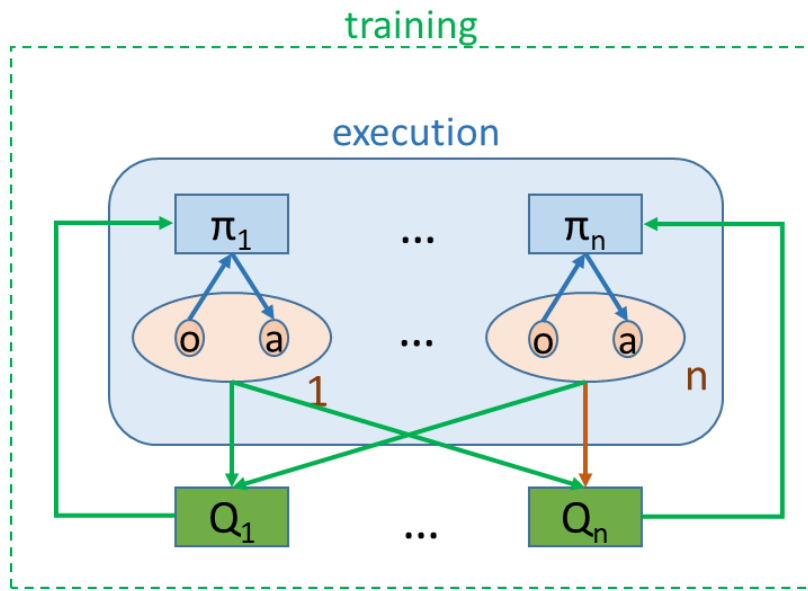

**Figure 7.** MADDPG interactions

The MADDPG algorithm (Lowe et al., 2017) builds upon the DDPG framework but utilizes a more centralized training, whereas the execution is still decentralized. In this setting, an agent (or actor) has access only to local information, whereas a centralized critic for each agent has access to global information and guides the actor at training time, like a coach. More specifically, MADDPG augments the critic network, a centralized action-value function, with extra information about the actions of all agents and the environment state information. Each critic outputs the Q-value for the corresponding agent. The actor policy ($\pi$), the critic network (Q), observations (o), actions (a), and their interactions are depicted in Figure 7. In general, the MADDPG algorithm: (1) leads to learned policies that only use local information at execution time (*i.e.,* their own observations), (2) does not assume a differentiable model of the environment dynamics or any particular structure on the communication method between agents, and (3) applies not only to cooperative interaction but to competitive or mixed interaction.

Models for the proof-of-concept room clearance scenario were trained both in Neural MMO and in RIDE using ML-Agents in conjunction with Shiva, which in turn used the MADDPG algorithm to discover behaviors for the red-team agents. In these experiments, red-team agents successfully discovered robust policies effective against the scripted behavior of the blue-team agents. For example, one particular layout (Figure 6) has two safe locations behind barricades in the middle of the room, and the agents can attack without receiving damage from these locations. For this layout, we have run experiments for a red team with two members and a blue team with three members. In this experiment, the red team started to consistently win in the scenario after 60,000 episodes, and after 75,000 episodes, the red-team started to win in the scenario without losing any of its members (Figure 8). As expected, the optimal policy here involves utilizing these locations to defeat the opposing team, and agents trained with MADDPG agents are able to discover that without explicit instructions about safe locations.

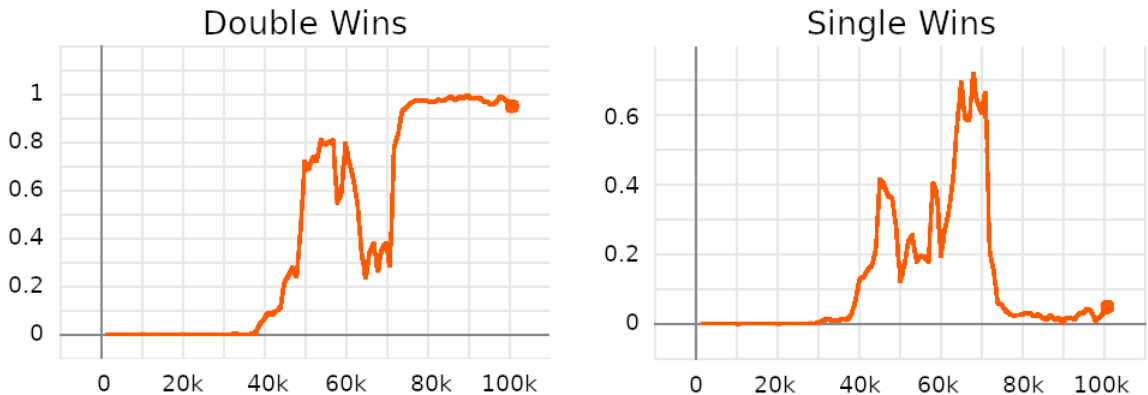

**Figure 7.** (a) Red team wins with both members standing    (b) Red team wins with only one member standing

**DISCUSSIONS**

The work with the proof-of-concept models shows excellent promise. Consequently, the focus now shifts applying MARL to more realistic scenarios devised on RIDE leveraging its capabilities. Furthermore, extending our framework by utilizing a capable cognitive architecture, Sigma, is an intriguing research direction. Current efforts building such a realistic scenario and potential paths for the integration of our framework with Sigma are discussed in the next two subsections.

**Scenario Generation**

The scenario being developed takes place at the Razish Military Training Village located in the National Training Center, Fort Irwin, CA. The terrain for this location is captured by the OWT project, which is currently available within the RIDE platform. This scenario is set up as a scout mission for the blue side located south of the village with the objective of occupying two key terrains (KT1 and KT2), the two hills overlooking the Razish Village as shown in Figure 9. The blue side comprises three teams: two fire teams and a heavy machine gun team. Each fire team has four members, whereas the heavy machine gun team has three members. The heavy machine gun team takes position at KT3, whereas the two fire teams head into the city using a bounding overwatch movement tactic. Here, it is assumed there is no aerial reconnaissance support, and the blue side needs to rely on line-of-sight while they are moving or engaging.

The red side comprises two small fire teams, each with two members. Their location is unknown to the blue side, but they are expected to be either at or around KT1 and KT2. There are a number of potential objectives for the red side: (1) Neutralize the blue side with minimal casualties, (2) Delay the entrance of the blue side for a certain amount of time, in order, for instance, to allow the arrival of supporting forces, and (3) Try to channelize the blue side to a specific area. The red side could have any of these objectives, and the objective of the red side is also unknown to the blue side.

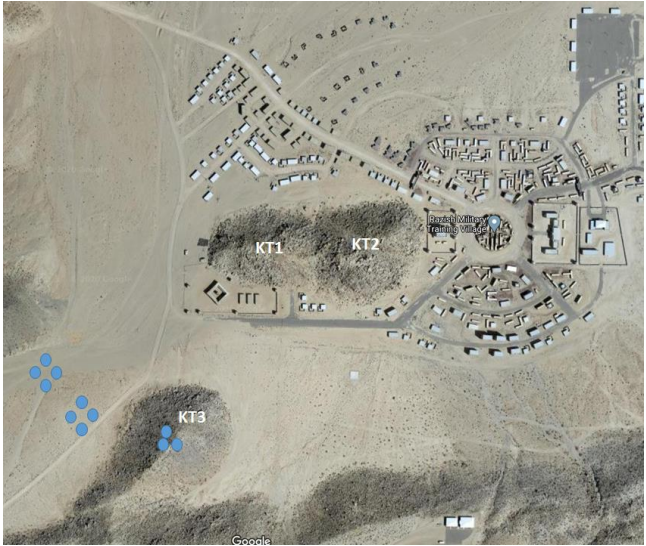

**Figure 9.** Scout mission at the Razish village

This scenario can be realistically played out in the RIDE simulation environment. The entities in RIDE can completely utilize the terrain, *e.g.,* climb hills, and use weapons with realistic attack behavior. Furthermore, RIDE provides accurate Line-of-Sight information based on environmental occlusion, providing a realistic framework to simulate the potential engagements between the blue and red sides.

Similar to our experiments with the proof-of-concept models, the initial focus here is to develop challenging red side behavior via leveraging multi-agent actor-critic models against the blue side, which follow a doctrine-based and pre-scripted behavior model. We work with several Subject Matter Experts (SMEs) to build up scripted behavior models for the blue side. These models will then be used in our MARL experiments leveraging RIDE to develop courses of actions for the red side, potentially varying based on the objective selected in a particular experiment. The generation of such courses of actions is valuable as it allows automated analysis of the most likely or most dangerous courses of actions that can be taken by the red side and identify potential vulnerabilities in the scripted behavior models for the blue side. As progress is made on the initial focus, we will explore learning effective behavior policies for the blue side as well since MARL allows agents on both teams – *e.g.,* blue and red side – to interact and adapt their respective policies. The concept of tournament play in MARL research (Jaderberg 2019), which uses gameplay among a diverse set of agents to achieve more effective behavior policies, could be of significant relevance here. This concept may yield a set of opponent behavior policies at different levels of difficulty.

**Sigma Integration**

Computational systems that can reason are cognitive systems, *e.g.,* humans, and can benefit from human-like social and cognitive capabilities. We previously had discussed that the synthetic characters in realistic simulations could benefit from being modeled as human-like social autonomous cognitive (HASC) systems (Ustun et al., 2018) that possess a *Theory-of-Mind* and that are perceptual, autonomous, interactive, affective, and adaptive. Such HASC systems, in general, require a computational intelligent behavior model that pushes the boundaries of how such a broad range of capabilities may be integrated. Such integration is an essential goal for a cognitive architecture (Kotseruba and Tsotsos 2018), which is a hypothesis concerning the fixed structures and mechanisms that, when supplemented by more variable knowledge and skills processed by the architecture, yields a cognitive system. Sigma (Σ) is a cognitive architecture that strives to combine what has been learned from independent work on symbolic cognitive architectures, probabilistic graphical models, and neural networks. Sigma leverages factor graphs (Kschischang, Frey & Loeliger, 2001) towards a uniform grand unification of symbolic, probabilistic, and neural reasoning for the construction of cognitive models that can possess symbolic, probabilistic, and neural reasoning (Rosenbloom, Demski, and Ustun 2017). In particular, Sigma leverages a generalization of factor graphs towards a uniform grand unification of not only traditional cognitive capabilities but also of critical non-cognitive aspects, creating unique opportunities for the construction of new kinds of cognitive models that possess a Theory of Mind and that are perceptual, autonomous, interactive, affective, and adaptive. Such a unification creates unique opportunities to augment the MARL models with symbolic and probabilistic reasoning capabilities to inject prior knowledge. For example, prior knowledge injection may help to devise MARL algorithms that consider doctrine and commander's guidance while generating courses of actions. Our current plans include to further explore the augmentation of MARL algorithms with the support of a cognitive architecture like Sigma to move towards having HASC systems as synthetic characters in our experiments.

## CONCLUSIONS

MARL models have been successfully applied to a variety of environments for the evaluation of competitive and collaborative dynamics between agents. Military training simulations can benefit from such models, but they can be complicated, and most of the existing simulation environments are not adequate to run RL experiments. In this paper, we have presented a realistic and fast simulation environment, RIDE, and a capable MARL framework, Shiva, to run RL experiments for military training scenarios. We have successfully created proof-of-concept combat behavior models leveraging our framework on realistic terrain as an essential step towards bringing ML into military training simulations. Furthermore, we introduced a squad vs. squad scenario taking place around the Razish Military Village at the National Training Center for MARL experiments. We also discussed how experimentations with such a scenario could be enriched by augmenting the models of the participating agents with cognitive capabilities along with symbolic and probabilistic reasoning via leveraging the Sigma cognitive architecture.

## ACKNOWLEDGMENTS

This effort has been sponsored by the U.S. Army. Statements and opinions expressed do not necessarily reflect the position or the policy of the United States Government, and no official endorsement should be inferred. We would also like to thank Ezequiel Donovan and Jorge Martinez for their contributions to the Shiva framework development.